\documentclass[11pt,a4paper]{article}
\usepackage[hyperref]{naaclhlt2018}
\usepackage{times}

\usepackage[pdftex]{graphicx}
\usepackage{verbatim}
\usepackage{amsmath}
\usepackage{enumitem}

\title{Content-Based Citation Recommendation}

\author{First Author \\
  Affiliation / Address line 1 \\
  Affiliation / Address line 2 \\
  Affiliation / Address line 3 \\
  {\tt email@domain} \\\And
  Second Author \\
  Affiliation / Address line 1 \\
  Affiliation / Address line 2 \\
  Affiliation / Address line 3 \\
  {\tt email@domain} \\}

\date{}

\author{Chandra Bhagavatula \\
  Allen Institute for AI \\
  {\tt chandrab@allenai.org} \\\And
  Sergey Feldman \\
  Data Cowboys \thanks{Work done while on contract with AI2} \\
  {\tt sergey@data-cowboys.com} \\\AND
  Russell Power \\
  Independent Researcher \thanks{Work done while at AI2} \\
  {\tt russell.power@gmail.com}  \\\And
  Waleed Ammar \\
  Allen Institute for AI \\
  {\tt waleeda@allenai.org} 
  }

\aclfinalcopy
\newcommand{\nnrank}{\texttt{NNRank}}
\newcommand{\nnselect}{\texttt{NNSelect}}
\newcommand{\corpus}{\texttt{OpenCorpus}}
\newcommand{\qdoc}{$d_q$}

\usepackage{array}
\newcolumntype{P}[1]{>{\centering\arraybackslash}p{#1}}

\hypersetup{draft}
\begin{document}

\maketitle

\begin{abstract}
We present a content-based method for recommending citations in an academic paper draft. We embed a given query document into a vector space, then use its nearest neighbors as candidates, and rerank the candidates using a discriminative model trained to distinguish between observed and unobserved citations. 
Unlike previous work, our method does not require metadata such as author names which can be missing, e.g., during the peer review process.
Without using metadata, our method outperforms the best reported results on PubMed and DBLP datasets with relative improvements of over 18\% in F1@20 and over 22\% in MRR. 
We show empirically that, although adding metadata improves the performance on standard metrics, it favors self-citations which are less useful in a citation recommendation setup.
We release an online portal for citation recommendation based on our method,\footnote{\url{http://labs.semanticscholar.org/citeomatic/}} and a new dataset \corpus{} of 7 million research articles to facilitate future research on this task.

\end{abstract}

\section{Introduction} \label{sec:intro}
Due to the rapid growth of the scientific literature, conducting a comprehensive literature review has become challenging, despite major advances in digital libraries and information retrieval systems.
Citation recommendation can help improve the quality and efficiency of this process by suggesting published scientific documents as likely citations for a query document, e.g., a paper draft to be submitted for ACL 2018. 
Existing citation recommendation systems rely on various information of the query documents such as author names and publication venue \cite{ren2014cluscite,Yu2012CitationPI}, or a partial list of citations provided by the author \cite{mcnee2002recommending,Liu2015ContextBasedCF,Jia2017AnAO} which may not be available, e.g., during the peer review process or in the early stage of a research project. 

Our method uses a neural model to embed all available documents into a vector space by encoding the textual content of each document.
We then select the nearest neighbors of a query document as candidates and rerank the candidates using a second model trained to discriminate between observed and unobserved citations. 
Unlike previous work, we can embed new documents in the same vector space used to identify candidate citations based on their text content, obviating the need to re-train the models to include new published papers. 
Further, unlike prior work \cite{yang2015network, ren2014cluscite}, our model is computationally efficient and scalable during both training and test time.

We assess the feasibility of recommending citations when some metadata for the query document is missing, and find that we are able to outperform the best reported results on two datasets while only using papers' textual content (i.e., title and abstract). 
While adding metadata helps further improve the performance of our method on standard metrics, we found that it introduces a bias for self-citation which might not be desirable in a citation recommendation system. 
See \S\ref{sec:experiments} for details of our experimental results.

\begin{figure*}[ht!]
    \centering
    \includegraphics[scale=0.57]{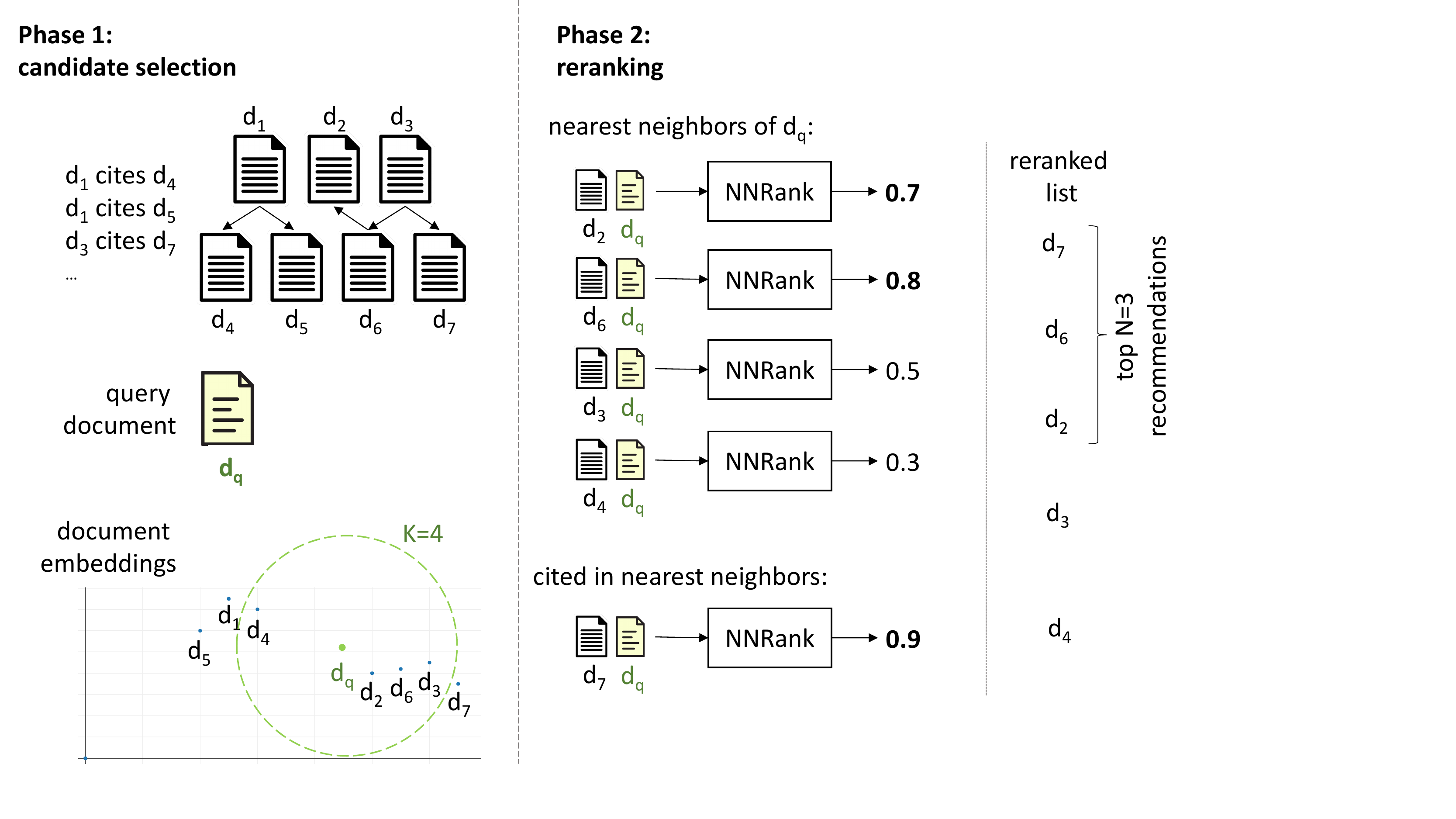}
    \vspace{-30pt}
    \caption{An overview of our Citation Recommendation system.
    In Phase 1 (\nnselect), we project all documents in the corpus (7 in this toy example) in addition to the query document $d_q$ into a vector space,
    and use its (K=4) nearest neighbors: $d_2, d_6, d_3,$ and $d_4$ as candidates. 
    We also add $d_7$ as a candidate because it was cited in $d_3$.
    In Phase 2 (\nnrank), we score each pair $(d_q,d_2), (d_q,d_6), (d_q,d_3), (d_q,d_4),$ and $(d_q, d_7)$ separately to rerank the candidates and return the top 3 candidates: $d_7$, $d_6$ and $d_2$.
    }
    \label{fig:example}
\end{figure*}

Our main contributions are: 
\begin{itemize}[noitemsep,nolistsep]
\item a content-based method for citation recommendation which remains robust when metadata are missing for query documents,
\item large improvements over state of the art results on two citation recommendation datasets despite omitting the metadata,
\item a new dataset of seven million research papers, addressing some of the limitations in previous datasets used for citation recommendation, and 
\item a scalable web-based literature review tool based on this work.\footnote{\url{https://github.com/allenai/citeomatic}}
\end{itemize}

\section{Overview} \label{sec:problem}\label{sec:overview}

We formulate citation recommendation as a ranking problem. 
Given a query document \qdoc{} and a large corpus of published documents, the task is to rank documents which should be referenced in \qdoc{}  higher than other documents.
Following previous work on citation recommendation, we use standard metrics (precision, recall, F-measure and mean reciprocal rank) to evaluate our predictions against gold references provided by the authors of query documents.

Since the number of published documents in the corpus can be large, it is computationally expensive to score each document as a candidate reference with respect to \qdoc.
Instead, we recommend citations in two phases: 
(i) a fast, recall-oriented candidate selection phase, and 
(ii) a feature rich, precision-oriented reranking phase. 
Figure \ref{fig:example} provides an overview of the two phases using a toy example.

\paragraph{Phase 1 - Candidate Selection:}
In this phase, our goal is to identify a set of candidate references for \qdoc{} for further analysis without explicitly iterating over all documents in the corpus.\footnote{In order to increase the chances that all references are present in the list of candidates, the number of candidates must be significantly larger than the total number of citations of a document, but also significantly smaller than the number of documents in the corpus.}
Using a trained neural network, we first project all published documents into a vector space such that a document tends to be close to its references.
Since the projection of a document is independent of the query document, the entire corpus needs to be embedded only once and can be reused for subsequent queries.
Then, we project each query document \qdoc{} to the same vector space and identify its nearest neighbors as candidate references.
See \S\ref{sec:candidate} for more details about candidate selection.

\paragraph{Phase 2 - Reranking:}
Phase 1 yields a manageable number of candidates making it feasible to score each candidate $d_{i}$ by feeding the pair $(d_q,d_i)$ into another neural network trained to discriminate between observed and unobserved citation pairs.
The candidate documents are sorted by their estimated probability of being cited in $d_q$, and top candidates are returned as recommended citations. 
See \S\ref{sec:sys} for more details about the reranking model and inference in the candidate selection phase.

\begin{figure*}[ht!]
    \centering
    \includegraphics[scale=0.55]{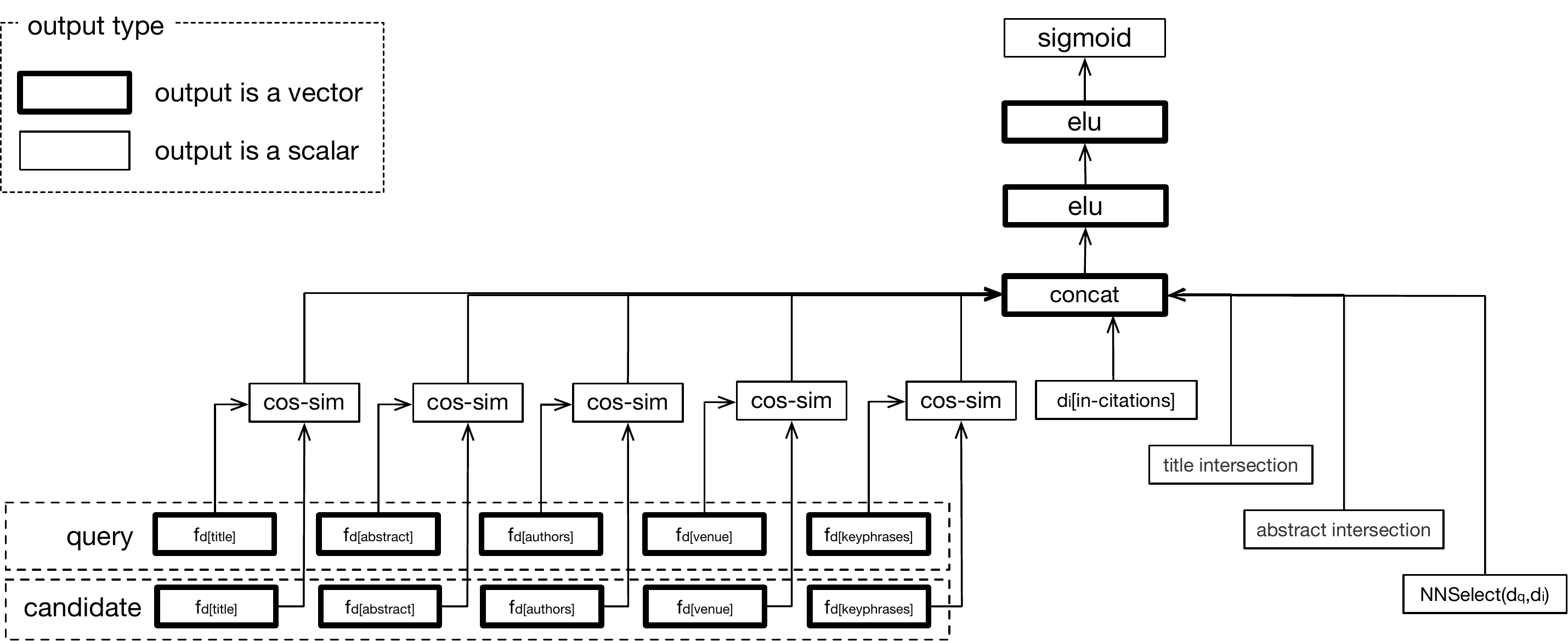}
    \caption{\nnrank{} architecture. For each of the textual and categorical fields, we compute the cosine similarity between the embedding for $d_q$ and the corresponding embedding for $d_i$.
Then, we concatenate the cosine similarity scores, the numeric features and the summed weights of the intersection words, followed by two dense layers with ELU non-linearities 
The output layer is a dense layer with sigmoid non-linearity, which estimates the probability that $d_q$ cites $d_i$.}
    \label{fig:citeomatic}
\end{figure*}

\section{Phase 1: Candidate Selection (\nnselect)} \label{sec:candidate}
In this phase, we select a pool of candidate citations for a given query document to be reranked in the next phase.
First, we compute a dense embedding of the query document $d_q$ using the document embedding model (described next), and select $K$ nearest neighbor documents in the vector space as candidates.\footnote{We tune $K$ as a hyperparameter of our method.}
Following \newcite{Strohman2007RecommendingCF}, we also include the outgoing citations of the $K$ nearest neighbors as candidates.

The output of this phase is a list of candidate documents $d_i$ and their corresponding scores $\nnselect(d_q,d_i)$, defined as the cosine similarity between $d_q$ and $d_i$ in the document embedding space.

\paragraph{Document embedding model.}
We use a supervised neural model to project any document $d$ to a dense embedding based on its textual content.
We use a bag-of-word representation of each textual field, e.g., $d[\text{title}] = \{ \text{`content-based', `citation', `recommendation'} \}$, and compute the feature vector:
\begin{align}
\mathbf{f}_{d[\text{title}]} &=  \sum_{t \in d[\text{title}]} w^{\text{mag}}_t \frac{\mathbf{w}^{\text{dir}}_t}{\| \mathbf{w}^\text{dir}_t\|_2}, \label{eq:textual_field}
\end{align}
where $\mathbf{w}^\text{dir}_t$ is a dense \underline{dir}ection embedding and $w^\text{mag}_t$ is a scalar \underline{mag}nitude for word type $t$.\footnote{The magnitude-direction representation is based on \newcite{SalimansK16} and was found to improve results in preliminary experiments, compared to the standard ``direction-only'' word representation.}
We then normalize the representation of each field and compute a weighted average of fields to get the document embedding, $\mathbf{e}_{d}$.
In our experiments, we use the title and abstract fields of a document $d$:
\begin{align*}
&\mathbf{e}_{d} =
\lambda^{\text{title}}  \frac{\mathbf{f}_{d[\text{title}]}}{\| \mathbf{f}_{d[\text{title}]}\|_2} + \lambda^{\text{abstract}}  \frac{\mathbf{f}_{d[\text{abstract}]}}{\| \mathbf{f}_{d[\text{abstract}]}\|_2}, 
\end{align*}
where $\lambda^{\text{title}}$ and $\lambda^{\text{abstract}}$ are scalar model parameters.

\paragraph{Training.}
We learn the parameters of the document embedding model (i.e., $\lambda^*, w^{\text{mag}}_*, \mathbf{w}^{\text{dir}}_*$) using a training set $\mathcal{T}$ of triplets $\langle d_q, d^+, d^- \rangle$ where $d_q$ is a query document, $d^+$ is a document cited in $d_q$, and $d^-$ is a document not cited in $d_q$.
The model is trained to predict a high cosine similarity for the pair $(d_q, d^+)$ and a low cosine similarity for the pair $(d_q, d^-)$ using the per-instance triplet loss \cite{Wang2014Triplet}:
\begin{align}
\text{loss}=\max\big(\alpha + s(d_q, d^-)  - s(d_q, d^+), 0\big),
\label{eq:loss}
\end{align}
where $s(d_i,d_j)$ is defined as the cosine similarity between document embeddings $\text{cos-sim}(\mathbf{e}_{d_i}, \mathbf{e}_{d_j})$. 
We tune the margin $\alpha$ as a hyperparameter of the model.
Next, we describe how negative examples are selected.

\paragraph{Selecting negative examples.}
Defining positive examples is straight-forward; we use any $(d_q,d^+)$ pair where a document $d_q$ in the training set cites $d^+$. However, a careful choice of negative training examples is critical for model performance. 
We use three types of negative examples:
\begin{enumerate}[noitemsep,nolistsep]
\item \textbf{Random:} any document not cited by $d_q$.
\item \textbf{Negative nearest neighbors:} documents that are close to $d_q$ in the embedding space, but are not cited in it.\footnote{Since the set of approximate neighbors depend on model parameters, we recompute a map from each query document to its $K$ nearest neighbors before each epoch while training the document embedding model.}
\item \textbf{Citation-of-citation:} documents referenced in positive citations of $d_q$, but are not cited directly in $d_q$. 
\end{enumerate}
In Appendix \S\ref{sec:hyperparams}, we describe the number of negative examples of  each type used for training. Next, we describe how to rerank the candidate documents.

\section{Phase 2: Reranking Candidates (\nnrank)} \label{sec:sys} \label{ref:doc_rep} \label{sec:architecture}
In this phase, we train another model which takes as input a pair of documents ($d_q$, $d_i$) and estimates the probability that $d_i$ should be cited in $d_q$.

\paragraph{Input features.}
A key point of this work is to assess the feasibility of recommending citations without using metadata, but we describe all features here for completeness and defer this discussion to  \S\ref{sec:experiments}.
For each document, we compute dense feature vectors $\mathbf{f}_{d[\text{field}]}$ as defined in Eq. \ref{eq:textual_field} for the following fields: title, abstract, authors, venue and keyphrases (if available).
For the title and abstract, we identify the subset of word types which appear in both documents (intersection), and compute the sum of their scalar weights as an additional feature, e.g., $\sum_{t\in \cap \text{title}} w^\cap_t$.
We also use \textit{log} number of times the candidate document $d_i$ has been cited in the corpus, i.e., \textit{log}($d_i[\text{in-citations}]$). 
Finally, we use the cosine similarity between $d_q$ and $d_i$ in the embedding space, i.e., $\text{cos-sim}(\mathbf{e}_{d_q},\mathbf{e}_{d_i})$. 

\paragraph{Model architecture.}
We illustrate the \nnrank~ model architecture in Figure \ref{fig:citeomatic}.
The output layer is defined as:
\begin{align}
s(d_i, d_j) &= \text{FeedForward}(\mathbf{h}), \label{eq:nnrank_loss} \\
\mathbf{h} &= \Big[ \mathbf{g}_{\text{title}}; \mathbf{g}_{\text{abstract}}; \mathbf{g}_{\text{authors}};  \mathbf{g}_{\text{venue}}; \nonumber  \\
&\hspace{20pt}  \mathbf{g}_{\text{keyphrases}}; \text{cos-sim}(\mathbf{e}_{d_q},\mathbf{e}_{d_i}); \nonumber \\
&\hspace{20pt} \scriptstyle\sum_{t \in \cap_{\text{title}}} w^\cap_t; \scriptstyle\sum_{t \in \cap_{\text{abstract}}} w^\cap_t;  \nonumber \\
&\hspace{20pt} d_i[\text{in-citations}] \Big], \nonumber  \\
\mathbf{g}_{\text{field}} &= \text{cos-sim}(\mathbf{f}_{d_q[\text{field}]}, \mathbf{f}_{d_i[\text{field}]}), \nonumber 
\end{align}
where `FeedForward' is a three layer feed-forward neural network with two exponential linear unit layers \cite{clevert2015fast} and one sigmoid layer.

\paragraph{Training.}
The parameters of the \nnrank{} model are $w^\text{mag}_*, \mathbf{w}^\text{dir}_*, w^\cap_*$ and parameters of the three dense layers in `FeedForward'. 
We reuse the triplet loss in Eq. \ref{eq:loss} to learn these parameters, but redefine the similarity function $s(d_i, d_j)$ as the sigmoid output described in Eq. \ref{eq:nnrank_loss}.

At test time, we use this model to recommend candidates $d_i$ with the highest $s(d_q,d_i)$ scores. 

\section{Experiments} \label{sec:experiments}
In this section, we describe experimental results of our citation recommendation method and compare it to previous work.

\paragraph{Datasets.} 
We use the DBLP and PubMed datasets \cite{ren2014cluscite} to compare with previous work on citation recommendation.
The DBLP dataset contains over 50K scientific articles in the computer science domain, with an average of 5 citations per article.
The PubMed dataset contains over 45K scientific articles in the medical domains, with an average of 17 citations per article. 
In both datasets, a document is accompanied by a title, an abstract, a venue, authors, citations and keyphrases. We replicate the experimental setup of \newcite{ren2014cluscite} 
by excluding papers with fewer than 10 citations and using the standard train, dev and test splits.

We also introduce \corpus,\footnote{\url{http://labs.semanticscholar.org/corpus/}} a new dataset of 7 million scientific articles primarily drawn from the computer science and neuroscience domain.
Due to licensing constraints, documents in the corpus do not include the full text of the scientific articles, but include the title, abstract, year, author, venue, keyphrases and citation information. 
The mutually exclusive training, development, and test splits were selected such that no document in the development or test set has a publication year less than that of any document in the training set.
Papers with zero citations were removed from the development and test sets.
We describe the key characteristics of \corpus{} in  Table \ref{tab:stats}.

\begin{table}[h]
\centering
\footnotesize
\begin{tabular}{l|r}
\textbf{Statistic}   &   \textbf{Value} \\ \hline
\# of documents in corpus  &    6.9 million \\
\# of unique authors   &   8.3 million \\
\# of unique keyphrases    &   823,677 \\
\# of unique venues    &   23,672 \\
avg. \# of incoming citations & 7.4 ($\pm$ 38.1) \\ 
avg. \# of outgoing citations & 8.4 ($\pm$ 14.4) \\ \hline
size of training set [years 1991 to 2014] & 5.5 million \\
size of dev set [years 2014 to 2015] & 689,000 \\
size of test set [years 2015 to 2016] & 20,000 \\
\end{tabular}
\caption{Characteristics of the \corpus{}. \label{tab:stats}}
\end{table}

\begin{table*}[ht!]
\centering
\tabcolsep=0.11cm
\begin{tabular}{l|ll|ll|ll}
\textbf{Method}   &   \multicolumn{2}{c|}{\textbf{DBLP}}    &    \multicolumn{2}{c|}{\textbf{PubMed}} &    \multicolumn{2}{c}{\textbf{\corpus{}}}\\ 
   &   F1@20   &   MRR   &    F1@20   &   MRR  & F1@20   &   MRR \\ \hline
BM25   & 0.119 & 0.425  &    0.209   &   0.574 & 0.058 & 0.218 \\ 
ClusCite     &    0.237    &    0.548    &     0.274    &    0.578  & -- & -- \\ \hline 
  \nnselect{}  & 0.282$\pm$0.002  & 0.579$\pm$0.007   & 0.309$\pm$0.001  &  0.699$\pm$0.001 & 0.109 & 0.221 \\ 
 \hspace{0.2cm} + \nnrank{}  &   0.302$\pm$0.001   &   0.672$\pm$0.015    &  0.325$\pm$0.001  &   0.754$\pm$0.003  & \textbf{0.126} & 0.330  \\ 
  \hspace{0.6cm}+ metadata  & \textbf{0.303}$\pm$0.001  & \textbf{0.689}$\pm$0.011   & \textbf{0.329}$\pm$0.001  & \textbf{0.771}$\pm$0.003  & 0.125 & \textbf{0.330} 
\end{tabular}
\caption{F1@20 and MRR results for two baselines and three variants of our method.
BM25 results are based on our implementation of this baseline, while ClusCite results are based on the results reported in \newcite{ren2014cluscite}.
``\nnselect'' ranks candidates using cosine similarity between the query and candidate documents in the embedding space (phase 1).
``\nnselect{} + \nnrank{}'' uses the discriminative reranking model to rerank candidates (phase 2), without encoding any of the metadata features.
``+ metadata'' encodes the metadata features (i.e., keyphrases, venues and authors), achieving the best results on all datasets. Mean and standard deviations are reported based on five trials.}
\label{tab:cluscite}
\end{table*}

\paragraph{Baselines.}
We compare our method to two baseline methods for recommending citations: ClusCite and BM25.
ClusCite \cite{ren2014cluscite} clusters nodes in a heterogeneous graph of terms, authors and venues in order to find related documents which should be cited.
We use the ClusCite results as reported in \newcite{ren2014cluscite}, which compared it to several other citation recommendation methods and found that it obtains state of the art results on the PubMed and DBLP datasets.
The BM25 results are based on our implementation of the popular ranking function Okapi BM25 used in many information retrieval systems. See Appendix \S\ref{appendix:bm25} for details of our BM25 implementation.

\paragraph{Evaluation.}
We use Mean Reciprocal Rank (MRR) and F1@20 to report the main results in this section.
In Appendix \S\ref{sec:detailed_results}, we also report additional metrics (e.g., precision and recall at 20) which have been used in previous work.
We compute F1@20 as the harmonic mean of the corpus-level precision and recall at 20 (P@20 and R@20).
Following \cite{ren2014cluscite}, precision and recall at 20 are first computed for each query document then averaged over query documents in the test set to compute the corpus-level P@20 and R@20. 

\paragraph{Configurations.}
To find candidates in \nnselect{}, we use the approximate nearest neighbor search algorithm \texttt{Annoy}\footnote{\url{https://github.com/spotify/annoy}}, which builds a binary-tree structure that enables searching for nearest neighbors in $O$(log \textit{n}) time. 
To build this tree, points in a high-dimensional space are split by choosing random hyperplanes. We use 100 trees in our approximate nearest neighbors index, and retrieve documents using the cosine distance metric. 

We use the \texttt{hyperopt} library\footnote{\url{https://github.com/hyperopt/hyperopt}} to optimize various hyperparameters of our method such as size of hidden layers, regularization strength and learning rate. 
To ensure reproducibility, we provide a detailed description of the parameters used in both \nnselect{} and \nnrank{} models, our hyperparameter optimization method and parameter values chosen in Appendix \S\ref{sec:hyperparams}. 

\paragraph{Main results.}
Table \ref{tab:cluscite} reports the F1@20 and MRR results for the two baselines and three variants of our method. Since the \corpus{} dataset is much bigger, we were not able to train the ClusCite baseline for it. \newcite{Totti2016AQA} have also found it difficult to scale up ClusCite to larger datasets.
Where available, we report the mean $\pm$ standard deviation based on five trials.

The first variant, labeled ``\nnselect,'' only uses the candidate selection part of our method  (i.e., phase 1) to rank candidates by their cosine similarity to the query document in the embedding space as illustrated in Fig.~\ref{fig:example}.
Although the document embedding space was designed to efficiently select candidates for further processing in phase 2, 
recommending citations directly based on the cosine distance in this space outperforms both baselines.

The second variant, labeled ``\nnselect{} + \nnrank{},'' uses the discriminative model (i.e., phase 2) to rerank candidates selected by \nnselect, without encoding metadata (venues, authors, keyphrases).
Both the first and second variants show that improved modeling of paper text can significantly outperform previous methods for citation recommendation, without using metadata.

The third variant, labeled ``\nnselect{} + \nnrank{} + metadata,'' further encodes the metadata features in the reranking model, and gives the best overall results.
On both the DBLP and PubMed datasets, we obtain relative improvements over 20\% (for F1@20) and 25\% (for MRR) compared to the best reported results of ClusCite.

In the rest of this section, we describe controlled experiments aimed at analyzing different aspects of our proposed method.

\paragraph{Choice of negative samples.}
As discussed in \S\ref{sec:candidate}, we use different types of negative samples to train our models.
We experimented with using only a subset of the types, while controlling for the total number of negative samples used, and found that using negative nearest neighbors while training the models is particularly important for the method to work.
As illustrated in Table \ref{tab:ablation}, on the PubMed dataset, adding negative nearest neighbors while training the models improves the F1@20 score from 0.306 to 0.329, and improves the MRR score from 0.705 to 0.771.
Intuitively, using nearest neighbor negative examples focuses training on the harder cases on which the model is more likely to make mistakes.

\begin{table}[ht]
\tabcolsep=0.11cm
\begin{tabular}{l|cc|cc}
   &   F1@20 & $\Delta$   &   MRR & $\Delta$  \\ \hline
Full model &   0.329  &  & 0.771  & \\  \hline
without intersection & 0.296 & \textbf{0.033} & 0.653 & \textbf{0.118} \\
without -ve NNs  &  0.306 &  0.016 &   0.705 & 0.066 \\
without numerical    &  0.314   & 0.008 &  0.735 & 0.036 \\
\end{tabular}
\caption{Comparison of PubMed results of the full model with model without (i) intersection features, (ii) negative nearest neighbors in training samples, and (iii) numerical features.}
\label{tab:ablation}
\end{table}

\paragraph{Valuable features.}
We experimented with different subsets of the optional features used in \nnrank~ in order to evaluate the contribution of various features.
We found intersection features, \nnselect~ scores, and the number of incoming citations to be the most valuable feature.
As illustrated in Table \ref{tab:ablation}, the intersection features improves the F1@20 score from 0.296 to 0.329, and the MRR score from 0.653 to 0.771, on the PubMed dataset.
The numerical features (\nnselect~ score and incoming citations) improve the F1@20 score from 0.314 to 0.329, and improves the MRR score from 0.735 to 0.771.
This shows that, in some applications, feeding engineered features to neural networks can be an effective strategy to improve their performance.

\paragraph{Encoding textual features.}
We also experimented with using recurrent and convolutional neural network to encode the textual fields of query and candidate documents, instead of using a weighted sum as described in Eq. \ref{eq:textual_field}.
We found that recurrent and convolutional encoders are much slower, and did not observe a significant improvement in the overall performance as measured by the F1@20 and MRR metrics. 
This result is consistent with previous studies on other tasks, e.g., \newcite{Iyyer2015DeepUC}.

\paragraph{Number of nearest neighbors.}
As discussed in \S\ref{sec:candidate}, the candidate selection step is crucial for the scalability of our method because it reduces the number of computationally expensive pairwise comparisons with the query document at runtime. 
We did a controlled experiment on the \corpus{} dataset (largest among the three datasets) to measure the effect of using different numbers of nearest neighbors, and found that both P@20 and R@20 metrics are maximized when \nnselect{} fetches five nearest neighbors using the approximate nearest neighbors index (and their out-going citations), as illustrated in Table \ref{tab:breakdown}.

\begin{table}[h!]
\centering
\begin{tabular}{c|c|c|c}
\textbf{\# of neighbors} & \textbf{R@20}   &   \textbf{P@20}  & \textbf{Time(ms)} \\ \hline
1  & 0.123 &  0.079  &   131 \\
\textbf{5} & \textbf{0.142}  &  \textbf{0.080}   &   144 \\
10 & 0.138 &  0.069   &  200 \\
50 & 0.081 &  0.040  &  362 \\
\end{tabular}
\caption{\corpus{} results for \nnselect{} step with varying number of nearest neighbors on 1,000 validation documents.}
\label{tab:breakdown}
\end{table}

\paragraph{Self-citation bias.}
We hypothesized that a model trained with the metadata (e.g., authors) could be biased towards self-citations and other well-cited authors. To verify this hypothesis, we compared two \nnrank{} models -- one with metadata, and one without. We measured the mean and max rank of predictions that had at least one author in common with the query document. This experiment was performed with the \corpus{} dataset.

\begin{figure}[ht!]
    \centering
    \includegraphics[scale=0.3]{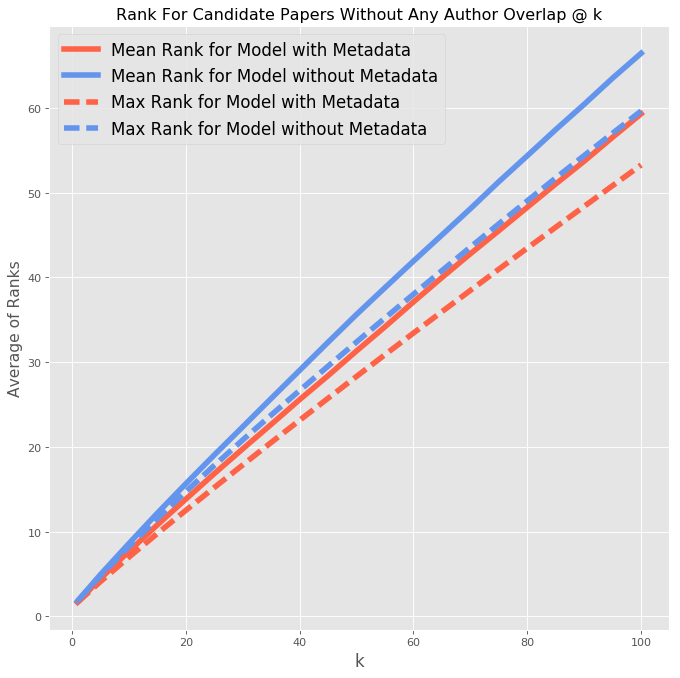}
    \caption{Mean and Max Rank of predictions with varying number of candidates.}
    \label{fig:bias-plot}
\end{figure}

A lower mean rank for \nnrank{} + Metadata indicates that the model trained with metadata tends to favor documents authored by one of the query document's authors. 
We verified the prevalence of this bias by varying the number of predictions for each model from 1 to 100. 
Figure \ref{fig:bias-plot} shows that the mean and max rank of the model trained with metadata is always lower than those for the model that does not use metadata. 

\section{Related Work}

Citation recommendation systems can be divided into two categories -- \textit{local} and \textit{global}. A local citation recommendation system takes a few sentences (and an optional placeholder for the candidate citation) as input and recommends citations based on the local context of the input sentences \cite{huang2015neural,he2010context, Tang2009ADA,huang2012recommending,He2011CitationRW}. A global citation recommendation system takes the entire scholarly article as input and recommends citations for the paper \cite{mcnee2002recommending,Strohman2007RecommendingCF,Nallapati2008JointLT,Kataria2010,ren2014cluscite}. We address the global citation recommendation problem in this paper.

A key difference of our proposed method compared to previous work is that our method is content-based and works well even in the absence of metadata (e.g. authors, venues, key phrases, seed list of citations). Many citation recommendation systems crucially rely on a query document's metadata. 
For example, the collaborative filtering based algorithms of \citet{mcnee2002recommending, Jia2017AnAO, Liu2015ContextBasedCF} require seed citations for a query document. \cite{ren2014cluscite,Yu2012CitationPI} require authors, venues and key terms of the query documents to infer interest groups and to extract features based on paths in a heterogeneous graph. In contrast, our model performs well solely based on the textual content of the query document. 

Some previous work (e.g. \cite{ren2014cluscite,Yu2012CitationPI}) have addressed the citation recommendation problem using graph-based methods. But, training graph-based citation recommendation models has been found to be expensive. For example, the training complexity of the ClusCite algorithm  \cite{ren2014cluscite} is cubic in the number of edges in the graph of authors, venues and terms. This can be prohibitively expensive for datasets as large as \corpus{}. On the other hand our model is a neural network trained via batched stochastic gradient descent that scales very well to large datasets \cite{bottou2010large}.

Another crucial difference between our approach and some prior work in citation prediction is that we build up a document representation using its constituent words only. Prior algorithms \cite{huang2015neural,huang2012recommending,Nallapati2008JointLT, Tanner2015AHG} learn an explicit representation for each training document separately that isn't a deterministic function of the document's words. This makes the model effectively transductive since a never-before-seen document does not have a ready-made representation. Similarly, \citet{huang2012recommending}'s method needs a candidate document to have at least one in-coming citation to be eligible for citation -- this disadvantages newly published documents. \citet{Liu2015ContextBasedCF} form document representations using citation relations, which are not available for unfinished or new documents. In contrast, our method does not need to be re-trained as the corpus of potential candidates grows. As long as the new documents are in the same domain as that of the model's training documents, they can simply be added to the corpus and are immediately available as candidates for future queries.

While the citation recommendation task has attracted a lot of research interest, a recent survey paper \cite{beel2016paper} has found three main concerns with existing work: (i) limitations in evaluation due to strongly pruned datasets, (ii) lack of details for re-implementation, and (iii) variations in performance across datasets. For example, the average number of citations per document in the DBLP dataset is 5, but \citet{ren2014cluscite} filtered out documents with fewer than 10 citations from the test set. This drastically reduced the size of the test set. We address these concerns by releasing a new large scale dataset for future citation recommendation systems. In our experiments on the \corpus{} dataset, we only prune documents with zero outgoing citations. We provide extensive details of our system (see Appendix \S\ref{sec:hyperparams}) to facilitate reproducibility and release our code\footnote{\url{https://github.com/allenai/citeomatic}}. We also show in experiments that our method consistently outperforms previous systems on multiple datasets.

Finally, recent work has combined graph node representations and text-based document representations using CCA \cite{Gupta2017ScientificAR}. This sort of approach can enhance our text-based document representations if a technique to create graph node representations at test-time is available.

\section{Conclusion}
In this paper, we present a content-based citation recommendation method which remains robust when metadata is missing for query documents, enabling researchers to do an effective literature search early in their research cycle or during the peer review process, among other scenarios. 
We show that our method obtains state of the art results on two citation recommendation datasets, even without the use of metadata available to the baseline method. 
We make our system publicly accessible online. 
We also introduce a new dataset of seven million scientific articles to facilitate future research on this problem.

\section*{Acknowledgements}
We would like to thank Oren Etzioni, Luke Zettlemoyer, Doug Downey and Iz Beltagy for participating in discussions and for providing helpful comments on the paper draft; Hsu Han and rest of the Semantic Scholar team at AI2 for creating the OpenCorpus dataset. We also thank Xiang Ren for providing the data used in their experiments on the DBLP and Pubmed datasets. Finally, we thank the anonymous reviewers for insightful comments on the draft.

\bibliography{naaclhlt2018}
\bibliographystyle{acl_natbib}

\newpage







\appendix

\section{Hyperparameter Settings} \label{sec:hyperparams}
Neural networks are complex and have a large number of hyperparameters to tune. This makes it challenging to reproduce experimental results. Here, we provide details of how the hyperparameters of the \nnselect{} and \nnrank{} models were chosen or otherwise set. We chose a subset of hyperparameters for tuning, and left the rest at manually set default values. Due to limited computational resources, we were only able to perform hyperparameter tuning on the development split of the smaller DBLP and Pubmed datasets. 

For DBLP and PubMed, we first ran Hyperopt\footnote{https://github.com/hyperopt/hyperopt} with 75 trials. Each trial was run for five epochs of 500,000 triplets each. The ten top performing of these models were trained for a full 50 epochs, and the best performing model's hyperparameters are selected. Hyperparameters for \nnselect{} were optimized for Recall@20 and those for the \nnrank{} model were optimized for F1@20 on the development set. The selected values for DBLP are reported in Table \ref{tab:hyper_dblp} and for PubMed are reported in Table \ref{tab:hyper_pubmed}. 

\corpus{} hyperparameters were set via informal hand-tuning, and the results are in Table \ref{tab:opencorpus_hyperparams}. A few miscellaneous parameters (not tuned) that are necessary for reproducibility are in Table \ref{tab:dataset_parameters}.

We briefly clarify the meaning of some parameters below:
\begin{itemize}
\item \textbf{Margin Multiplier} - The triplet loss has variable margins for the three types of negatives: $0.1\gamma, 0.2\gamma, \text{and } 0.3\gamma$. We treat $\gamma$ as a hyperparameter and refer to it as the margin multiplier.
\item \textbf{Use Siamese Embeddings} - For the majority of our experiments, we use a \emph{Siamese} model \cite{Bromley93}. That is, the textual embeddings for the query text and abstract share the same weights. However, we had a significantly larger amount of data to train \nnrank{} on \corpus{}, and found that non-Siamese embeddings are beneficial.
\item \textbf{Use Pretrained} - We estimate word embeddings on the titles and abstracts of \corpus{} using Word2Vec implemented by the gensim Python package\footnote{https://radimrehurek.com/gensim/}.
\end{itemize}

\section{Margin Loss Details}
When computing the margins for the triplet loss, we use a boosting function for highly cited documents. The full triplet loss function is as follows:
\begin{align*}
\text{max} \big(
	&\gamma\alpha(d^-) \\ 
    &+ s(d_q, d^-) + B(d^-) \\
    &- s(d_q, d^+) - B(d^+) \\
    &, 0\big)
\end{align*}
where $\gamma$ is the margin multiplier, and $\alpha(d^-)$ varies based on the type of negative document:
\begin{itemize}[leftmargin=*]
\item $\alpha(d^-) = 0.3$ for random negatives
\item $\alpha(d^-) = 0.2$ for nearest neighbor negatives
\item $\alpha(d^-) = 0.1$ for citation-of-citation negatives.
\end{itemize}
The boosting function is defined as follows:
$$B(d) = \frac{\sigma\left(\frac{d[\text{in-citations}]}{100} \right)}{50}$$
where $\sigma$ is the sigmoid function and $d[\text{in-citations}]$ is the number of times document $d$ was cited in the corpus. The boosting function allows the model to slightly prefer candidates that are cited more frequently, and the constants were set without optimization.

\section{Nearest Neighbors for Training Details}
When obtaining nearest neighbors for negative examples during training, we use a heuristic to find a subset of the fetched nearest neighbors that are sufficiently wrong. That is, these are non-citation samples that look dissimilar in the original text but similar in the embedding space. This procedure is as follows for each training query:
\begin{enumerate}
\item Compute the Jaccard similarities between a training query and all of its true citations using the concatenation of title and abstract texts.
\item Compute the bottom fifth percentile Jaccard similarity value. I.e. the value below which only the bottom 5\% most least textually similar true citations fall. For example, if the Jaccard similarities range from 0.2 to 0.9, the fifth percentile might plausibly be 0.3.
\item Use the Annoy index computed at the end of the previous epoch to fetch nearest neighbors for the query document.
\item Compute the textual Jaccard similarity between all of the nearest neighbors and the query document.
\item Retain nearest neighbors that have a smaller Jaccard similarity than the fifth percentile. Using the previous example, retain the nearest neighbors that have a lower Jaccard similarity than 0.3.
\end{enumerate}

\section{BM25 Details}
\label{appendix:bm25}
\begin{table}[ht!]
\small
\centering
\begin{tabular}{p{2.5cm}|cc|cc}
\textbf{BM25 Implementation}   &   \multicolumn{2}{c|}{\textbf{DBLP}}    &    \multicolumn{2}{c}{\textbf{PubMed}} \\
   &   F@20   &   MRR   &   F@20   &   MRR \\ \hline
\citet{ren2014cluscite}   &   0.111   &   0.411   &    0.153   &   0.497   \\
Our approximation   &   0.119   &    0.425   &   0.209   &   0.574   \\
\end{tabular}
\caption{Results of our BM25 implementation on DBLP and Pubmed datasets.}
\label{table:bm25}
\end{table}

Okapi-BM25 is a popular ranking function. We use BM25 as an IR-based baseline for the task of citation recommendation. For the DBLP and Pubmed datasets, BM25 performance is provided in \citet{ren2014cluscite}. To create a competitive BM25 baseline for \corpus{}, we first created indexes for the DBLP and Pubmed datasets and tuned the query to approximate the performance reported in previous work. We used Whoosh\footnote{\url{https://pypi.python.org/pypi/Whoosh/}} to create an index. We extract the key terms (using Whoosh's \texttt{key\_terms\_from\_text method}) from the title and abstract of each query document. The key terms from the document are concatenated to form the query string. Table \ref{table:bm25} shows that our BM25 is a close approximation to the BM25 implementation of previous work and can be reliably used as a strong IR baseline for \corpus{}. In Table \ref{tab:cluscite}, we report results on all three datasets using our BM25 implementation.

\section{Key Phrases for \corpus}
In the \corpus{} dataset, some documents are accompanied by automatically extracted key phrases. Our implementation of automatic key phrase extraction is based on standard key phrase extraction systems -- e.g. \cite{caragea2014kp,caragea2014citation,lopez2010humb}. We first extract noun phrases using the Stanford CoreNLP package \cite{Manning2014TheSC} as candidate key phrases. Next, we extract corpus level and document level features (e.g. term frequency, document frequency, n-gram probability etc.) for each candidate key phrase.  Finally, we rank the candidate key phrases using a ranking model that is trained on author-provided key phrases as gold labels.

\begin{table*}[ht!]
\centering
\resizebox{\textwidth}{!}{%
\begin{tabular}{l|ll|ll}
\hline
 & \multicolumn{2}{c|}{\textbf{\nnselect}} & \multicolumn{2}{c}{\textbf{\nnrank}}   \\
\textbf{Hyperparameter} & \multicolumn{1}{c}{\textbf{Range}} & \multicolumn{1}{c|}{\textbf{Chosen Value}} & \multicolumn{1}{c}{\textbf{Range}}                 & \multicolumn{1}{c}{\textbf{Chosen Value}} \\ \hline
learning rate       & $[1\text{e-}5, 1\text{e-}4, \ldots, 1\text{e-}1]$    & $0.01$                               & $[1\text{e-}5, 1\text{e-}4, \ldots, 1\text{e-}1]$    & $0.01$                               \\
l2 regularization   & $[0, 1\text{e-}7, 1\text{e-}6, \ldots, 1\text{e-}2]$ & $0$                                  & $[0, 1\text{e-}7, 1\text{e-}6, \ldots, 1\text{e-}2]$ & $1\text{e-}3$                        \\
l1 regularization   & $[0, 1\text{e-}7, 1\text{e-}6, \ldots, 1\text{e-}2]$ & $1\text{e-}7$                        & $[0, 1\text{e-}7, 1\text{e-}6, \ldots, 1\text{e-}2]$ & $1\text{e-}4$                        \\
word dropout        & $[0, 0.05, 0.1, \ldots, 0.75]$                       & $0.60$                               & $[0, 0.05, 0.1, \ldots, 0.75]$                       & $0.35$                               \\
margin multiplier   & $[0.5, 0.75, 1.0, 1.25, 1.5]$                        & $1.0$                                & $[0.5, 0.75, 1.0, 1.25, 1.5]$                        & $0.5$                                \\
dense dimension     & $[25, 50, \ldots, 325]$                              & $300$                                & $[25, 50, \ldots, 325]$                              & $175$                                \\
metadata dimension      & -                                                  & -                                  & $[5, 10, \ldots, 55]$                                & $45$                                 \\
use pretrained      & {[}true, false{]}                                    & true                                 & {[}true, false{]}                                    & false                                \\
finetune pretrained & {[}true, false{]}                                    & true                                 & {[}true, false{]}                                    & -                                 \\ 
number ANN neighbors & - & 10 & - & - \\
triplets per batch size & - & $256$ & -  & $256$ \\
triplets per epoch & - & $500000$ & - & $500000$  \\
triplets per training & - & $2500000$ & - & $2500000$  \\
use Siamese embeddings & - & true & - & true\\
\hline
\end{tabular}%
}
\caption{DBLP hyperparameter tuning results. Note that the dense dimension when using pretrained vectors is fixed to be 300. A '-' indicates that the variable was not tuned. }
\label{tab:hyper_dblp}
\end{table*}

\begin{table*}[ht!]
\centering
\resizebox{\textwidth}{!}{%
\begin{tabular}{l|ll|ll}
\hline
 & \multicolumn{2}{c|}{\textbf{\nnselect}} & \multicolumn{2}{c}{\textbf{\nnrank}}   \\
\textbf{Hyperparameter} & \multicolumn{1}{c}{\textbf{Range}} & \multicolumn{1}{c|}{\textbf{Chosen Value}} & \multicolumn{1}{c}{\textbf{Range}}                 & \multicolumn{1}{c}{\textbf{Chosen Value}} \\ \hline
learning rate           & $[1\text{e-}5, 1\text{e-}4, \ldots, 1\text{e-}1]$    & $0.001$                              & $[1\text{e-}5, 1\text{e-}4, \ldots, 1\text{e-}1]$    & $0.001$                               \\
l2 regularization       & $[0, 1\text{e-}7, 1\text{e-}6, \ldots, 1\text{e-}2]$ & $0$                                  & $[0, 1\text{e-}7, 1\text{e-}6, \ldots, 1\text{e-}2]$ & $0$                                  \\
l1 regularization       & $[0, 1\text{e-}7, 1\text{e-}6, \ldots, 1\text{e-}2]$ & $1\text{e-}6$                        & $[0, 1\text{e-}7, 1\text{e-}6, \ldots, 1\text{e-}2]$ & $1\text{e-}6$                                  \\
word dropout            & $[0, 0.05, 0.1, \ldots, 0.75]$                       & $0.55$                               & $[0, 0.05, 0.1, \ldots, 0.75]$                       & $0.1$                               \\
margin multiplier       & $[0.5, 0.75, 1.0, 1.25, 1.5]$                        & $0.5$                                & $[0.5, 0.75, 1.0, 1.25, 1.5]$                        & $1.5$                                \\
dense dimension         & $[25, 50, \ldots, 325]$                              & $325$                                & $[25, 50, \ldots, 325]$                              & $150$                                \\
metadata dimension      & -                                                  & -                                  & $[5, 10, \ldots, 55]$                                & $40$                                 \\
use pretrained          & {[}true, false{]}                                    & false                                & {[}true, false{]}                                    & false                                 \\
finetune pretrained     & {[}true, false{]}                                    & -                                  & {[}true, false{]}                                    & - \\ 
number ANN neighbors & - & 10 & - & - \\
triplets per batch size & - & $256$ & -  & $256$ \\
triplets per epoch & - & $500000$ & - & $500000$  \\
triplets per training & - & $2500000$ & - & $2500000$  \\
use Siamese embeddings & - & true & - & true\\
\hline
\end{tabular}%
}
\caption{PubMed hyperparameter tuning results. Note that the dense dimension when using pretrained GloVe vectors is fixed to be 300. A '-' indicates that the variable was not tuned.}
\label{tab:hyper_pubmed}
\end{table*}

\begin{table*}[ht!]
\centering
\begin{tabular}{l|ll}
\hline
\textbf{Hyperparameter} & \textbf{PubMed/DBLP Value} & \textbf{\corpus{}  Value}\\ \hline
title/abstract vocabulary size & $200000$ & $200000$ \\
maximum title length & $50$ & $50$ \\
maximum abstract length & $500$ & $500$ \\
training triplets per query & $6$ & $6$  \\
min \# of papers per author included & 1 & 10 \\
min \# of papers per venue included & 1 & 10 \\
min \# of papers per keyphrases included & 5 & 10 \\
max authors per document & 8 & 8 \\
max keyphrases per document & 20 & 20 \\
minimum true citations per document & 2/1 & 2 \\
maximum true citations per document & 100 & 100 \\
optimizer & LazyAdamOptimizer* & Nadam** \\
use magnitude-direction embeddings & true & true \\
reduce learning rate upon plateau & false & true \\
\hline
\end{tabular}
\caption{Per-dataset parameters. These were hand-specified. *LazyAdamOptimizer is part of TensorFlow. **Nadam is part of Keras.}
\label{tab:dataset_parameters}
\end{table*}

\begin{table*}[ht!]
\centering
\begin{tabular}{l|ll}
\hline
\textbf{Hyperparameter} & \textbf{\nnselect{} Value} & \textbf{\nnrank{}  Value}\\ \hline
learning rate & $0.001$ & $0.001$ \\
l2 regularization & $1\text{e-}5$ & $1\text{e-}5$ \\
l1 regularization & $1\text{e-}7$ & $1\text{e-}7$ \\
word dropout & 0.1 & 0.1 \\
margin multiplier & $1.0$ & $1.0$ \\
dense dimension & $75$ & $75$ \\
metadata dimension & - & $25$ \\
use pretrained & false & false\\
number ANN neighbors & 5 & - \\
triplets per batch size & $256$ & $32$ \\
triplets per epoch & $2500000$ & $2500000$  \\
triplets per training & $25000000$ & $100000000$  \\
use Siamese embeddings & true & false\\ 
\hline
\end{tabular}
\caption{Hyperparameters used for \corpus{}}
\label{tab:opencorpus_hyperparams}
\end{table*}

\section{Detailed Experimental Results}
\label{sec:detailed_results}

Table \ref{tab:appendix-cluscite} compares \nnrank{} with previous work in detail on DBLP and Pubmed datasets. ClusCite \cite{ren2014cluscite} clusters nodes in a heterogeneous graph of terms, authors and venues in order to find related documents which should be cited. ClusCite obtains the previous best results on these two datasets. L2-LR \cite{Yu2012CitationPI} uses a linear combination of meta-path based linear features to classify candidate citations. We show that \nnrank{} (with and without metadata) consistently outperforms ClusCite and other baselines on all metrics on both datasets.

\begin{table*}[ht!]
\resizebox{\textwidth}{!}{%
\begin{tabular}{l|ccccc|ccccc}
\hline
 &   \multicolumn{5}{c|}{\textbf{DBLP}}    &    \multicolumn{5}{c}{\textbf{PubMed}} \\
 \textbf{Method}    &   \textbf{P@10}   &   \textbf{P@20}   &    \textbf{R@20}   &   \textbf{F1@20}   &   \textbf{MRR}   &   \textbf{P@10}   &   \textbf{P@20}   &    \textbf{R@20}   &   \textbf{F1@20}   &   \textbf{MRR} \\ \hline
BM25   &   0.126   &   0.0902   &   0.1431   &   0.11 & 0.4107   &   0.1847   &    0.1349   & 0.1754   &   0.15   &   0.4971  \\ 
L2-LR & 0.2274 &  0.1677 & 0.2471 & 0.200 & 0.4866 & 0.2527  & 0.1959 & 0.2504 & 0.2200 & 0.5308 \\ 
ClusCite   &   0.2429   &   0.1958    &    0.2993    &  0.237    &    0.5481    &    0.3019    &    0.2434    &  0.3129  &    0.274    &    0.5787 \\ \hline
\nnselect{}   & 0.287   &   0.230  & 0.363  & 0.282  & 0.579   &  0.388  & 0.316  & 0.302  & 0.309  &  0.699 \\ 
\hspace{0.2cm} + \nnrank{}  &  0.339   &  0.247   &   0.390   &   0.302   &   0.672   &   0.421   &   0.332   &  0.318   &   0.325   &   0.754   \\ 
  \hspace{0.6cm}+ metadata  & \textbf{0.345}   &   \textbf{0.247}  & \textbf{0.390}  & \textbf{0.303}  & \textbf{0.689}   & \textbf{0.429} & \textbf{0.337}  & \textbf{0.322}  & \textbf{0.329}  & \textbf{0.771} \\ \hline
\end{tabular}
}
\caption{Comparing \nnrank{} with ClusCite.  \cite{ren2014cluscite} have presented results on several other topic-based, link-based and network-based citation recommendation methods as baselines. For succinctness, we show results for the best system, Cluscite, and two baselines BM25 and L2-LR.}
\label{tab:appendix-cluscite}
\end{table*}


\end{document}